\documentclass{article}
\usepackage{ijcai22}

\usepackage{times}
\usepackage{soul}
\usepackage{url}

\usepackage[hidelinks,backref=page]{hyperref}
\renewcommand*\backref[1]{\ifx#1\relax \else (cit. on p. #1) \fi}

\usepackage[utf8]{inputenc}
\usepackage[small]{caption}
\usepackage{graphicx}
\usepackage{amsmath}
\usepackage{amsthm}
\usepackage{amsfonts}
\usepackage{mathrsfs}
\usepackage{booktabs}
\usepackage{algorithm}
\usepackage{algorithmic}
\usepackage{wrapfig}
\usepackage{pdfpages}
\usepackage{tikz}
\usetikzlibrary{shapes.geometric, matrix,backgrounds, positioning}
\usepackage{pgfplots}
\usepackage{rotating}
\pgfplotsset{compat=1.17}

\pgfdeclarelayer{background}
\pgfdeclarelayer{foreground}
\pgfsetlayers{background,main,foreground}

\usepackage{xcolor}

\graphicspath{{../images/}}

\title{Modeling Oceanic Variables with Dynamic Graph Neural Networks}

\author{
Caio F. D. Netto\footnote{Equal contribution by first and second authors.
All other authors made significant suggestions and the last five authors
oversaw the whole project and secured funding.}$^1$
\and
Marcel R. de Barros$^{\ast1}$\and
Jefferson F. Coelho$^1$\and
Lucas P. de Freitas$^1$\and\\
Felipe M. Moreno$^{1,3}$\and
Marlon S. Mathias$^{3, 4}$\and
Marcelo Dottori$^2$\and \\
Fábio G. Cozman$^{1, 3}$\and
Anna H. R. Costa$^{1, 3}$\and
Edson S. Gomi$^{1, 3}$\and
Eduardo A. Tannuri$^{1, 3}$
\affiliations
$^1$Escola Politécnica -- University of Sao Paulo, Brazil\\
$^2$Instituto Oceanográfico -- University of Sao Paulo, Brazil\\
$^3$Center for Artificial Intelligence (C4AI) -- University of Sao Paulo, Brazil\\
$^4$Instituto de Estudos Avançados (IEA) -- University of Sao Paulo, Brazil\\
\emails
\{caio.netto, marcel.barros, jfialho, lfreitasp2001, felipe.marino.moreno,
marlon.mathias, mdottori, fgcozman, anna.reali, gomi, eduat\}@usp.br
}

\begin{document}

\maketitle

\begin{abstract}
Researchers typically resort to numerical methods to understand and predict
ocean dynamics, a key task in mastering environmental phenomena.
Such methods may not be suitable in scenarios where the topographic map is complex,   knowledge about the underlying processes is incomplete, or the application is time critical. On the other hand, if  ocean dynamics are observed, they can be exploited by recent machine learning methods. In this paper we describe a  data-driven method to predict environmental variables such as current velocity and sea surface height in the region of Santos-Sao Vicente-Bertioga Estuarine System in the southeastern coast of Brazil. 
Our model exploits both temporal and spatial inductive biases by joining state-of-the-art sequence models (LSTM and Transformers) and relational models (Graph Neural Networks) in an end-to-end   framework that learns both the temporal features and the spatial relationship shared among observation sites. We compare our results with the \textit{Santos Operational Forecasting System} (SOFS). Experiments show that better results are attained by our model, while maintaining flexibility and little domain knowledge dependency.
\end{abstract}

\section{Introduction}

Machine Learning (ML) has shown promising results in many fields, not only as an applied tool but also as a main driver of scientific discovery \cite{Raccuglia2016,Cranmer2020,Jumper2021}. If in the past environmental models were governed by first-principle models based on established science, it is now clear that this is not the ideal approach in a handful of situations. There are domains in which the underlying phenomena and the governing equations are not well understood or the complexity of the problem is enormous, making it unfeasible in practice to solve them through first-principle models. 

One such domain is the oceanic one. Being able to understand and predict how the ocean dynamics works is a major concern for   coastal countries both economically and socially. Techniques have been proposed so as to model ocean dynamics numerically \cite{POM87,Ribeiro2019,CostaSOFS2020}. The goal is often to anticipate extreme phenomena, for instance  storm surges, which can cause transportation delays and accidents, when forecasting current velocity and sea surface height. However, as stated before, physical models are costly to design, implement, and maintain, given that they require accurate measurements of the region of interest topology for   spatial representations  and for boundaries conditions. 

Conversely, ML models have been successfully applied to a variety of   physical problems \cite{Tianfang15,Sanchez2020,Lira2022}  and in   particular in the oceanic domain \cite{Ibarra2015,xiaoSpatiotemporalDeepLearning2019,Netto2020}. These models circumvent the computational cost of physical models while attaining excellent results. As measurement tools have improved,   both public and private interest to record oceanic data has risen in recent years \cite{PIANC2012}. One can explore ML models with such historical data. aiming to implement and deploy better ocean dynamics prediction systems. However, as   sensors and measurement tools are ``in the wild'', the observation of hydrodynamic and meteorological variables are highly affected, for instance, by extreme or unexpected environmental conditions  or technical glitches. So, key variables for ocean dynamics, such as current and wind velocity and sea surface height, tend to display noisy and missing  data.

In order to build a data-driven model that deals with those constraints, specially missing data and diverse kinds of time series, a variety of ML models have been proposed. Sequence models like LSTM and Transformers \cite{Hochreiter97,Vaswani2017}  and Graph Neural Networks (GNN) \cite{scarselliGraphNeuralNetwork2009} are suitable for this task as they incorporate both temporal and spatial inductive biases into their architecture.

In this work we aim to address a time series forecasting problem in the context of ocean dynamics, proposing a spatio-temporal GNN architecture to predict current velocity and sea surface height, and using multivariate time series data collected at the Santos-Sao Vicente-Bertioga Estuarine System. Forecasting oceanic variables is a major concern, in general, for both public actors and private port authorities, and it is even more in that area, which is home to the largest port in Latin America, the Port of Santos.

We summarize the main contributions  of this work as follows: 
\begin{itemize}
    \item We propose a general model capable of dealing with problems that have both temporal and spatial dimensions and significant levels of missing data.
    \item We address a real problem in the context of physical sciences, using a data-driven method rather than physical model-based simulations.
    \item We demonstrate experimentally that the proposed model surpasses the physical model SOFS, our baseline, by $27$\% on water current speed prediction and $14$\% on  water current direction prediction, while maintaining excellent performance in forecasting sea surface height, considering the Willmott index \cite{Willmott1981}, also known as the Index of Agreement (IoA).
\end{itemize}
Regarding the structure of the paper, we introduce in Section~\ref{sec:related_works} previous works related to machine learning models in the context of relational learning, multivariate time series prediction and ocean domain. After, in Section~\ref{sec:background} we describe the basic concepts and the problem we address, while in Section~\ref{sec:model_ssvbes} we use that theory to model our problem. In Section~\ref{sec:architecture} we detail our architecture. Sections~\ref{sec:exp_setup} and \ref{sec:conclusions} close the paper reporting our experimental setup, the obtained results and discussions, conclusions and planned future work.

\section{Related Work} \label{sec:related_works}

In an effort to address all dimensions in which our problem is embedded, we quickly highlight the most recent research that develops new methods for forecasting multivariate time series in the oceanic domain, where temporal and spatial biases matter  and data-driven methods produce better results.
A number of statistical methods have been proposed to address these kinds of problems. \citeauthor{weiMultivariateTimeSeries2019} (\citeyear{weiMultivariateTimeSeries2019}) presents in an intuitive and comprehensive way a handful of those methods, like autoregressive integrated moving-average for the case of multivariate time series data, applying them, for example, to real public health problems. Nonetheless, these statistical models are known to face difficulties in capturing long-term relations and seasonal components. 

Due to their flexibility, ML models have shown excellent results in multivariate time series forecasting. For example, simple models like Quantile Regression Forests (QRF) were used in \cite{Moreno2022} to mitigate the error of a physics-based numerical model, built to forecast the surface current of an ocean region. The authors proposed a different and flexible framework that models the problem from the ``backdoor'': they exploit the existence of a physical model and build one that models the residual error between the former and the surface current true value.

Recent works have been focused on applying deep learning models or implementing their own, due to both the impressive performance in a variety of tasks and the capability of incorporating domain specific knowledge into those models' architecture. For instance, in \cite{ziat2017spatio} the authors proposed a spatio-temporal Recurrent Neural Network (RNN) model for time series forecasting, combining a sequential model with the dependence between time series of locations spatially separated. In the field of application of this work, others have even proposed the combination of state-of-the-art models like Convolutional Neural Networks (CNN) and RNN, building spatio-temporal time series models that benefit from both inductive biases \cite{xiaoSpatiotemporalDeepLearning2019}.

In order to better model relationships between entities of a problem, Graph Neural Networks (GNN) showed promising results and well-established theory to do so. \citeauthor{Lira2022} (\citeyear{Lira2022}) address the time series problem of frost forecasting, using GNNs with attention to model an experimental site spatially and temporally. In \cite{Cao2020}, the authors compose both convolution and sequential learning in a relational architecture (GNN), extracting richer features in the frequency domain through Fourier Transform, aiming to model multivariate time series forecasting problems. Specifically to the ocean domain, \cite{Netto2020} propose an application of GNNs to model ocean variables in the form of time series, in a real problem of an economically important coastal region, similarly to \cite{Lira2022}.
\section{Background} \label{sec:background}

In this section we outline the main ideas behind modeling time series as a graph and GNNs (Subsection \ref{subsec:gnn}), and the problem we address in this work (Subsection \ref{subsec:ssvbes_problem}).

\subsection{Dynamic Graph Neural Networks} \label{subsec:gnn}

Our work adopts   terminology by \cite{kazemiRepresentationLearningDynamic2020} in
which a \emph{continuous-time dynamic graph} (CTDG) is a pair
$(\mathcal{G},\mathcal{O})$ where $\mathcal{G}$ is a (static) initial graph and
$\mathcal{O}$ is a set of observations/events of the form $(event\ type,event,timestamp)$ that can alter graph structure,
node attributes, and edge attributes.

Each static graph is defined by a pair of sets $(\mathcal{V},\mathcal{E})$. $\mathcal{V}$ is a set of nodes and
$\mathcal{E}$ is a set of edges. Following the structure proposed by \cite{satorrasMultivariateTimeSeries2022}, we work with a fully connected graph.
A \emph{discrete-time dynamic graph} (DTDG) can be defined as a set
of snapshots $\{\mathcal{G}^{1},\mathcal{G}^{2},\ldots,\mathcal{G}^{T}\}$ sampled from an underlying CTDG.

Finally, a \emph{Graph Neural Network} \cite{scarselliGraphNeuralNetwork2009} can be described as message passing mechanism
which iteractively updates nodes' hidden representations. This mechanism can be summarized 
as follows:
\begin{align}
    \resizebox{0.85\hsize}{!}{$\mathbf{x}_i^{(k)} = \gamma^{(k)} \left( \mathbf{x}_i^{(k-1)}, \square_{j \in \mathcal{N}(i)} \, \phi^{(k)}\left(\mathbf{x}_i^{(k-1)}, \mathbf{x}_j^{(k-1)}\right) \right)$},
    \label{mpnn}
\end{align}
where $\mathbf{x}_i^{(k)}$ is the hidden representation (or node features) of a
node $i$ at iteration $k$, $\phi^{(k)}$ is a differentiable \emph{message function}
that constructs a message to be propagated
based on both sender ($\mathbf{x}_j$) and receiver ($\mathbf{x}_i$) nodes, $\square$ denotes a differentiable,
permutation invariant \emph{aggregation function} that aggregates all
messages from neighbour nodes and $\gamma^{(k)}$ is a differentiable \emph{update function},
that updates the node's hidden representation.

Both $\phi$ and $\gamma$ are parametrized by a set of learnable parameters that are
iteratively updated during the training process. Examples of such functions would be Multi Layer Perceptrons (MLPs).

\subsection{The SSVBES Forecasting Problem} \label{subsec:ssvbes_problem}

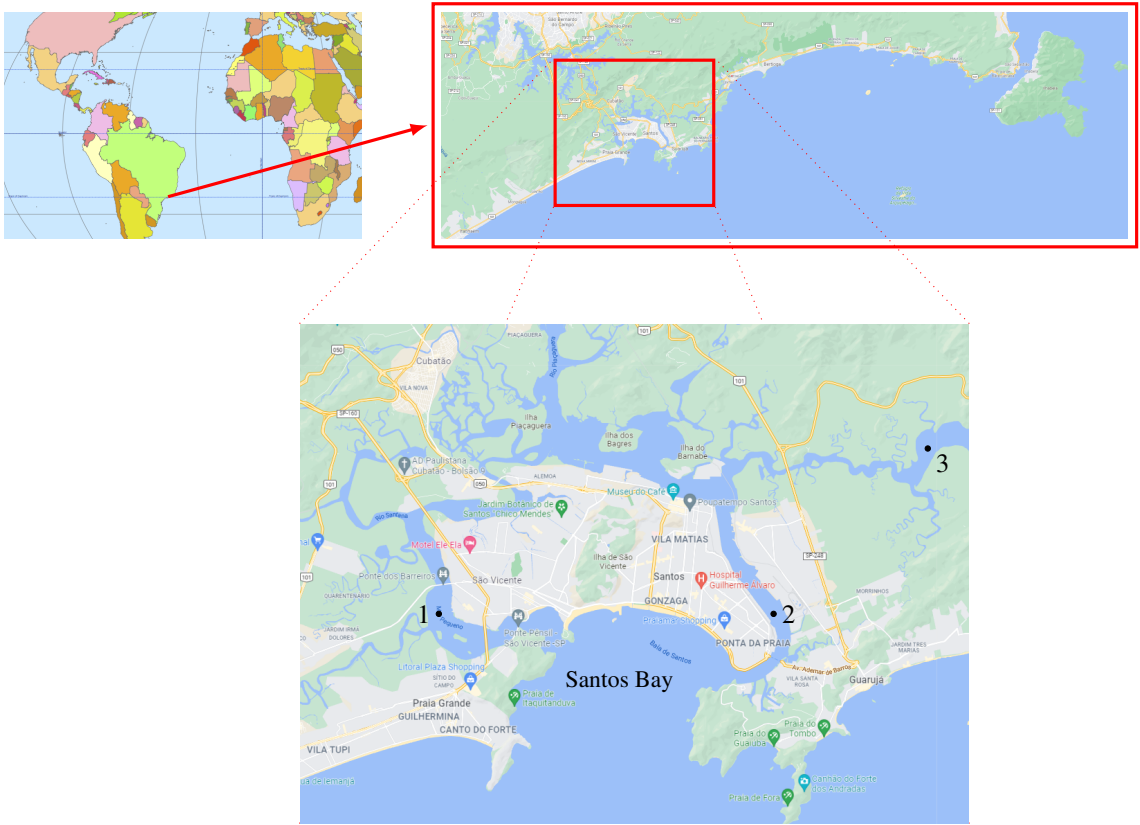
\begin{figure*}[!ht]
    \centering
    \scalebox{1.0}{
        \begin{tikzpicture}
            \node at (2,1) {\includegraphics[height=3cm]{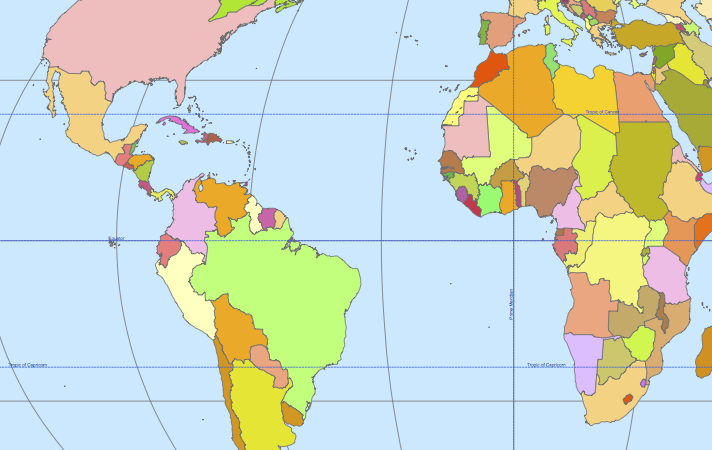}};
            \node[draw,red,very thick] (one) at (10,1) {\includegraphics[height=3cm]{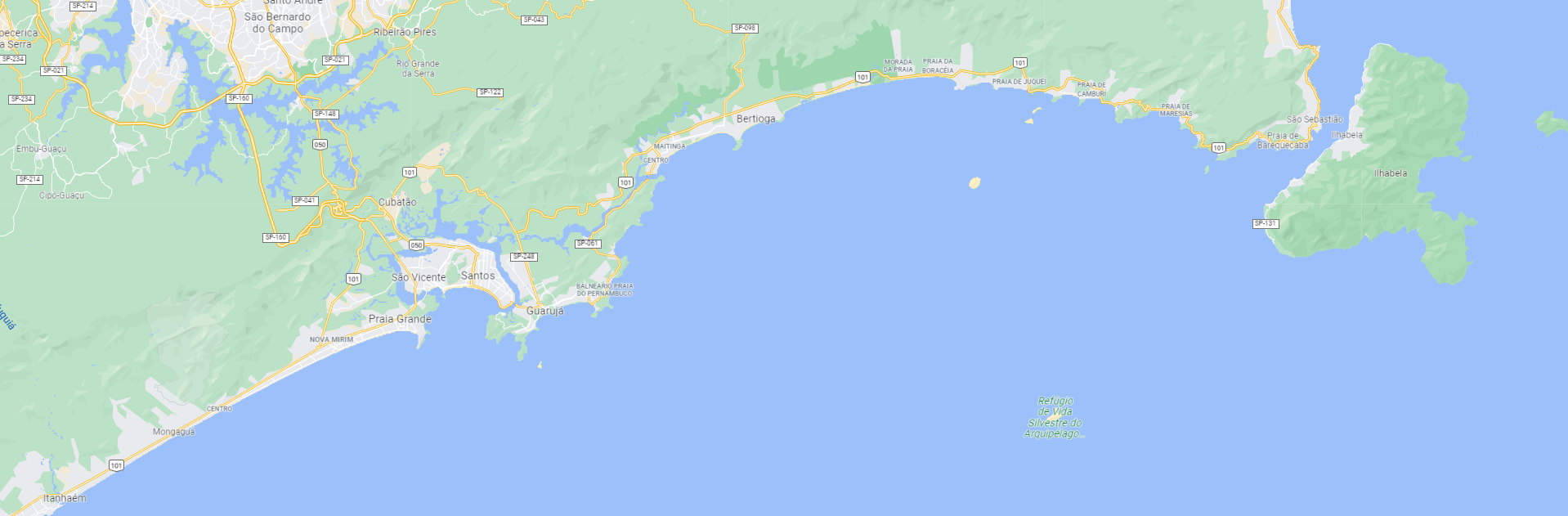}};
            \node[draw, red, very thick, minimum width=60, minimum height=55] (rect) at (8, 0.9) {};
    
            \begin{pgfonlayer}{foreground}
                \node[anchor=north, inner sep=0pt] (two) at ([xshift=-2cm, yshift=-1cm]one.south){\includegraphics[width=0.5\textwidth]{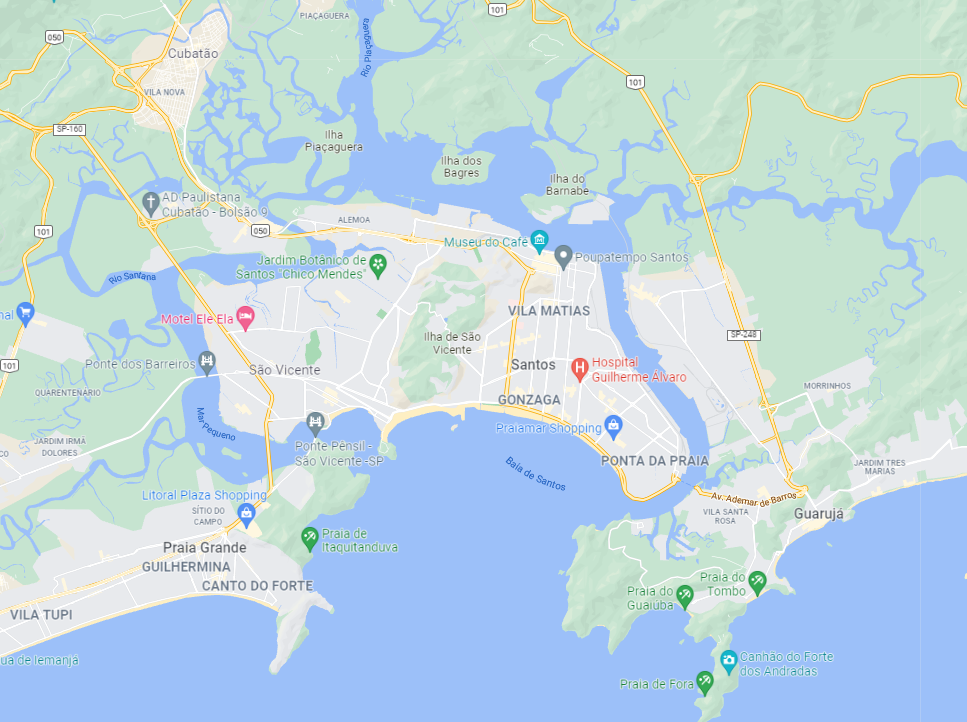}};
                \node at (7.8,-6.4) {\small Santos Bay};
                \draw[fill=black] (5.4, -5.5) circle (1pt);
                \draw[fill=black] (9.85,-5.5) circle (1pt);
                \draw[fill=black] (11.9,-3.3) circle (1pt);
                https://www.overleaf.com/project/6256df55c6b72880fa5d3fda
                \node at (5.2, -5.5) {1};
                \node at (10.05,-5.5) {2};
                \node at (12.1,-3.5) {3};
            \end{pgfonlayer}
    
            \draw[red, dotted] (two.north east) -- (rect.north east);
            \draw[red, dotted] (two.north west) -- (rect.north west);
            \draw[red, dotted] (two.south east) -- (rect.south east);
            \draw[red, dotted] (two.south west) -- (rect.south west);
            \draw[->,>=latex,very thick,red] (1.8,0.05)--(5.25,1);
        \end{tikzpicture}
    }
    \caption{The Santos-Sao Vicente-Bertioga Estuarine System. In the bottom figure,
    the main locations:
        1 - Sao Vicente Channel;
        2 - Santos Port Channel;
        3 - Bertioga Channel.
    }
    \label{fig:Santos}
\end{figure*}

The Santos-Sao Vicente-Bertioga Estuarine System (SSVBES) is located on the southeast coast of Brazil, as part of the South Brazil Bight. This estuarine system, like others over the world, has its hydrodynamics driven mainly by three forcing processes, which are the astronomical tide, meteorological tide and river discharge. The meteorological tide is dependent on the synoptic winds that blow over the adjacent continental shelf, associated with simple Ekman dynamics and, therefore, is a process that occurs off the Santos Bay and enters it as a gravity wave. When winds blow from North-Northeast, sea surface height (SSH) decreases and, when winds from South-Southwest are dominant, SSH increases. We present that region in Figure \ref{fig:Santos} as well as its three main channels, Sao Vicente Channel, Santos Port Channel, and Bertioga Channel, that constitute the estuarine system.

An effort to understand the SSVBES behavior under the action of these forcings has been coded into
the \textit{Santos Operational Forecasting System} (SOFS) \cite{CostaSOFS2020}, which provides daily forecasts for the region. SOFS adopts a finite difference model that uses the \textit{Navier-Stokes} equation with sigma vertical coordinates, is forced by the winds, tides, density gradients and river discharge, with good results for both SSH and currents.

However, data availability of river discharge limits the accuracy outputs of this model because this is not a variable easily obtained in a near-real time frame. This limitation is mainly observed in   current results, as river discharge in estuaries causes changes in   flow, directly, and in   vertical density stratification, which is also capable of changing currents.

\section{Modeling SSVBES with DTDGs} \label{sec:model_ssvbes}

\begin{figure}[!htb]
    \centering
    \begin{tikzpicture}
        \node[anchor=south west,inner sep=0] at (0,0) {\includegraphics[width=0.4\textwidth]{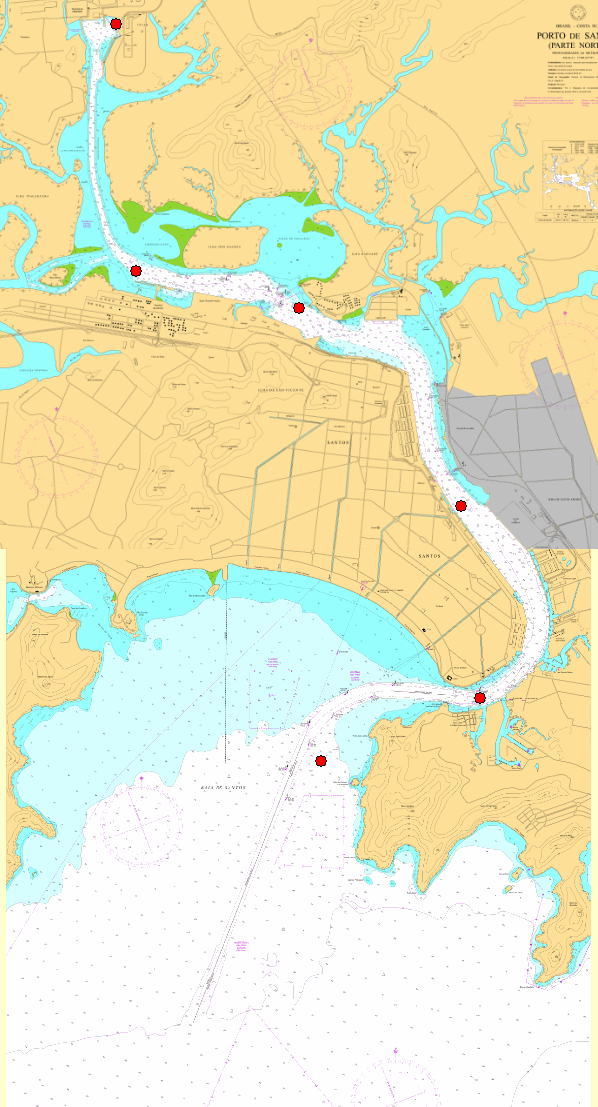}};
        \node at (1.7, 12.9) {1};
        \node at (1.63, 9.6) {2};
        \node at (3.56, 9.17) {3};
        \node at (5.2, 7.18) {4};
        \node at (5.716, 4.55) {5};
        \node at (3.55, 4.15) {6};
    \end{tikzpicture}
    \caption{Santos-Sao Vicente-Bertioga Estuarine System, and the location of all six observation sites used in the experiments:
        1) TIPLAM,
        2) Alemoa,
        3) Ilha Barnabé,
        4) CPSP,
        5) Praticagem,
        6) Palmas.
    }
    \label{fig:Sites}
\end{figure}

We model SSVBES as a DTDG comprised of snapshots sampled from an underlying CTDG. 
Each node in this CTDG is a pair
$(type,location)$ denoting a node type and a location. Possible node types are oceanic variables
such as SSH, water current, and wind. Each different $type$ has $k_{type}$
features of which $l_{type}$ are labels. Possible locations are
measuring stations in the SSVBES system from which data is
collected. For this region, we have access to observations in six different sites located as indicated in Figure \ref{fig:Sites}.

Each data point collected is an event
associated with the node $(type, location)$.
For example, a measurement from a water current velocity sensor in Praticagem station -- located in the observation site 5 --
is considered an event associated with the node $(current,praticagem)$.

\subsection{Sampling the CTDG}
To build a snapshot of the CTDG associated with an instant $t$ we use the following procedure:
\begin{enumerate}
    \item Define a past window time interval (\emph{past\_len}) and filter events
          having ${timestamp \in [{t-past\_len}, t[}$.
    \item Define a future/prediction window time interval (\emph{future\_len})
          and filter events having ${timestamp \in [{t}, {t+future\_len}[}$.
    \item A node $x_i$ associated with the pair $(type,location)$ is part of the snapshot
          if the past window has at least one event associated with that pair.
    \item Each node $x_i$ associated with the pair $(type,location)$ is built with
          three time-sorted sequences of events:
          \begin{itemize}
              \item $x_i^{(past)} \in \mathbb{R}^{ps_i \times k_{type}}$: sequence with $ps_i$ past events;
              \item $x_i^{(future)} \in \mathbb{R}^{fs_i \times (k_{type}-l_{type})}$: sequence with $fs_i$ future events without label columns;
              \item $y_i \in \mathbb{R}^{fs_i \times l_{type}}$: sequence with $fs_i$ future events with only label columns.
          \end{itemize}
\end{enumerate}

The future features sequence $x_i^{(future)}$ stores only features that are associated
to future events  but whose values are available beforehand such as
astronomical tide and timestamp encoding columns.

Note that $ps_i$ and $fs_i$ may vary for each node $x_i$ depending on its data
collection periodicity and eventual missing data points (e.g. sensor outage).

To build the complete DTDG graph $\mathcal{G}=\{\mathcal{G}^1,\mathcal{G}^2,\ldots,\mathcal{G}^N\}$ we sample $\boldsymbol{N}$ uniformly
spaced snapshots from the underlying CTDG.

\subsection{Encoder-Decoder modeling} \label{enc_dec_model}
Following the definition from \cite{kazemiRepresentationLearningDynamic2020},
an encoder receives a dynamic graph as input and outputs an embedding function,
while a decoder receives an embedding function as input and performs
a forecasting task.

Our forecast model is comprised of two sets: a set of encoders $Enc=\{{Enc}^{type_1},{Enc}^{type_2},\ldots,{Enc}^{type_E}\}$, and a set of decoders $Dec=\{{Dec}^{type_1},{Dec}^{type_2},\ldots,{Dec}^{type_E}\}$, where $E$ represents the number of oceanic variables being processed.

Each encoder $Enc_{type}$ is comprised of two modules, namely, a \emph{temporal encoder}
and a \emph{GNN} for information propagation between nodes.
While the temporal encoder is different for each $type$, the \emph{GNN}
module is shared amongst all nodes.
\paragraph*{The temporal encoder} receives $x_i^{(past)}$ and $x_i^{(future)}$
as input and outputs a single fixed size hidden representation $h_i$.
\paragraph*{The GNN} updates all $h_i$ based on graph neighbourhood as
detailed by (\ref*{mpnn}), enabling information from different nodes to be shared.
\paragraph*{The decoder} receives $h_i$ as input and outputs a forecast $\hat{y_i}$
for a subset of $\mathcal{V}$.

This setup enables, for example, the sea surface height and water current nodes to have
different \emph{temporal encoders} and \emph{decoders} that may
have different architectures.
This flexible approach can be adapted to a vastly different set of input variables,
as any time series can be encoded into a node representation and have its information
propagated through the graph structure.

\section{Proposed Architecture} \label{sec:architecture}

\begin{figure*}[htb]
    \begin{center}
        \includegraphics[scale=1.2]{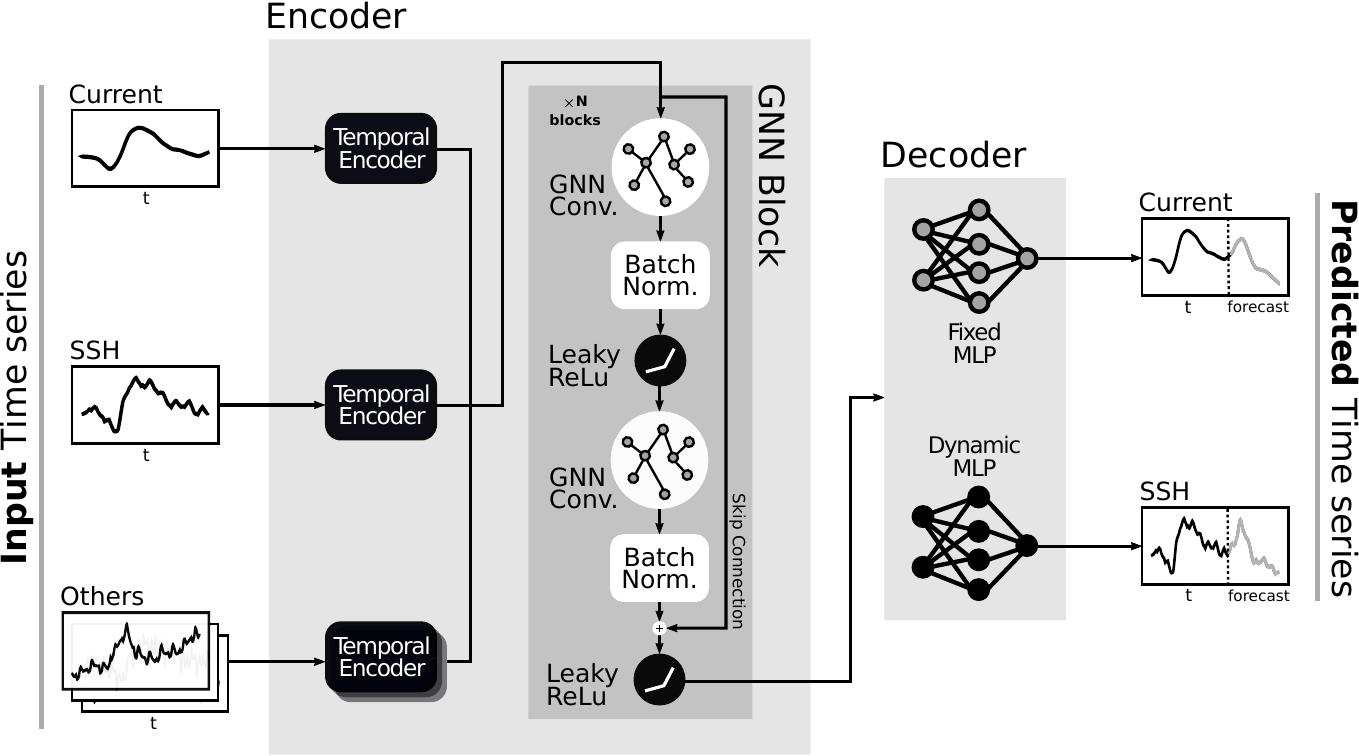}
        \caption{Overall model architecture indicating each $type$ associated to a \emph{temporal encoder} and a \emph{decoder} module, while all $types$ share the same GNN structure. Internally, the GNN is a sequence of GNN blocks with normalization layers, graph convolutions and nonlinearities.}
        \label{fig:arch}
    \end{center}
\end{figure*}

We propose a modularized architecture, shown in Figure \ref{fig:arch}, that fits in the encoder-decoder category, thus enabling the use of different sequence processing architectures for encoding and decoding of each different oceanic variable. Also, we separate the concept of a \emph{GNN Block} from its inner \emph{GNN Convolution} object allowing it to be replaced by newer graph convolution architectures while still maintaining normalization layers and skip connections. In this section we describe each one of these components.

\subsection{Encoders}

Each $Enc_{type}$ encoder receives two sequences of variable length and outputs a
single hidden representation $h_i$ of fixed length $embed\_size$.

\subsubsection{Temporal Encoding}

We implement the \emph{temporal encoder} described in Section \ref{enc_dec_model} with two different
sequence models, namely a Transformer and an LSTM, and compare results. 
Both models can be described as:
\begin{align*}
    \xi \colon \mathbb{R}^{{ps_i} \times {k_{type}}} \times \mathbb{R}^{{fs_i} \times {(k_{type}-l_{type})}} & \to \mathbb{R}^{embed\_size}.
\end{align*}
In both cases, we use two instances of the same architecture to encode both $x_i^{(past)}$ and $x_i^{(future)}$. To form the final
fixed size embedding, we concatenate the results
into a single vector $h_i = [h_i^{(past)}, h_i^{(future)}]$.

\subsubsection{Spatial Encoding}
Given the fixed sized embedding $h_i$ we use a GNN block to update
the embedding with information from incoming edges. The GNN block is described as:
\begin{align*}
    \mathbb{G} \colon \mathbb{R}^{{embed\_size}} & \to \mathbb{R}^{embed\_size}.
\end{align*}
Inspired by \cite{heDeepResidualLearning2016} we establish a block with a skip connection
and normalization layers as depicted in Figure \ref{fig:arch}.

Each GNN block contains graph convolutions that are responsible for aggregating neighbourhood
information. Our implementation uses a Graph Attention Convolution, first proposed by
\cite{velickovicGraphAttentionNetworks2018} and then further improved by
\cite{brody2021attentive}.

\subsection{Decoders}

Each $Dec_{type}$ decoder receives a
single hidden representation $h_i$ of fixed length $embed\_size$ as input and
outputs a sequences of length $fs_i$.
Our implementation uses two different decoder architectures. 
A \emph{Fixed Output Size MLP} is employed to decode water current velocity data of the form:
\begin{align*}
    \mathbb{D}_{fix} \colon \mathbb{R}^{{embed\_size}} & \to \mathbb{R}^{{max(fs_i)} \times l_{type}}.
\end{align*}
We call this architecture a fixed output size one, because $max(fs_i)$ is the maximum length of the output sequence and is defined beforehand based on training data.
To decode sea surface height related data, we employ a \emph{Dynamic Output Size MLP} that can be viewed as:
\begin{align*}
    \mathbb{D}_{dyn} \colon \mathbb{R}^{{embed\_size}+{embed\_size}/2} & \to \mathbb{R}^{l_{type}}.
\end{align*}
In this architecture, the decoder also receives $h_i^{(future)}$ as input and concatenates it to
the embedding $h_i$ to form the final input to the MLP.
This format resulted in much better results for SSH forecast.

We hypothesize that this
is due to the fact that $h_i^{(future)}$ for $type = ``SSH"$ contains astronomical
tide information which is the major
factor defining SSH behaviour. The model seems to benefit from receiving this information
directly in the decoder. We leave a more detailed study of this behaviour to future work.
\section{Experimental Setup} \label{sec:exp_setup}
Here we describe our datasets and its characteristics (Subsection \ref{subsec:datasets}), the model configurations and the stack of tools used to implement and run our experiments (Subsection \ref{subsec:model_configs}), and our experimental design (Subsections \ref{subsec:scenarios}  and \ref{subsec:discussion})

\subsection{Datasets}\label{subsec:datasets}

As input to the model, we used   environmental sensing data collected from the 6 regions of the SSVBES between 1/1/2019 and 1/9/2021 (totaling 974 days). 

\begin{figure}[!htb]
    \centering
    \includegraphics[width=0.48\textwidth]{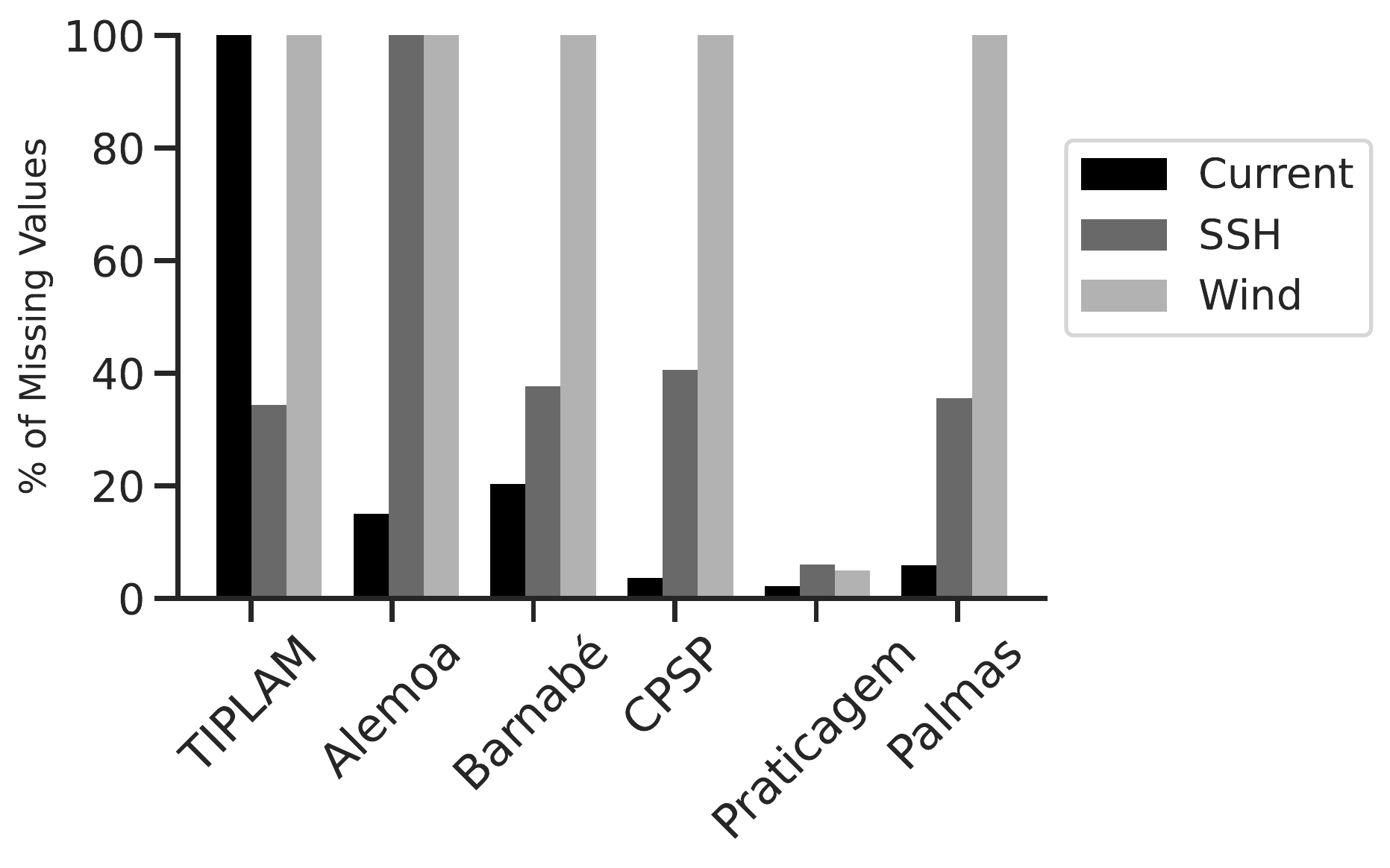}
    \caption{Distribution of missing values for different event types from the SSVBES observation sites.}
    \label{fig:dataset}
\end{figure}

As the sensors responsible for data acquisition are directly affected by climate and environmental conditions, the input features presents the following percentage of missing data:  Current: $24.3$\%, SSH: $42.1$\% and Wind: $84.1$\%, totaling approximately $50.1$\% missing data from the sensing date range.
The data are also unbalanced as to the percentage of data obtained in the different locations, as seen in Figure \ref{fig:dataset}.

The monitoring data was aggregated using a 30-minute step between windows,  which means there are 3 graphs $\mathcal{G}$ in the flow data for each hour. In addition, data inputs are normalized by Z-score.

To deal with the characteristics of the available data, the proposed graph model does not need strict  input shape or contiguous data. Each node of the model structure receives an independent data window as input after available features data encoding.

\subsection{Scenarios}\label{subsec:scenarios}
To assess our model performance, we designed 4 experimental scenarios.\\

\paragraph{Water Current and SSH Forecasting} In the first two experiments we evaluated the performance of the proposed model in predicting the current velocity and sea surface height, separately. We compared these results to SOFS. We used the same set of hyperparameters to train both models.
\paragraph{Fully Connected \textit{vs} Same-variable \textit{vs} Disconnected Graph} Next, we experimented and compared three graph topologies: a fully connected graph, a graph with connections only between nodes of the same \emph{type} and a fully disconnected graph. Analogously to the previous experiment, all hyperparameters were fixed and we only modified our adjacency matrix.
\paragraph{LSTM \textit{vs} Transformers as  Temporal Encoders} Lastly, given our framework flexibility, we compared different state-of-the-art sequence models as temporal encoders. For that experiment, we contrasted the results with an LSTM against Transformers as encoders, keeping the same set of hyperparameters for both cases.\\

We tested for each scenario a set of combinations of target variables, which are current velocity and SSH. In the first and second experiments (respectively, predicting current velocity and sea surface height), we fed as input to our model (i) only the corresponding target variable and (ii) both, to predict the corresponding target variable. In the last two experiments we used the 4 possible combinations of input variables. Table \ref{tab:exp_design} summarizes this combination for input and target variables.

\begin{table}[htb]
    \centering
    \resizebox{\columnwidth}{!}{%
        \begin{tabular}{*{11}{l}}
            \toprule
            Scenarios & \multicolumn{1}{c}{\textbf{Inputs}} & \multicolumn{1}{c}{\textbf{Targets}} \\
            \midrule
            $1^{st}$ Experiment & [Current; Current+SSH; Current+SSH+Wind] & [Current] \\
            $2^{nd}$ Experiment & [SSH; Current+SSH] & [SSH] \\
            $3^{rd}$ Experiment & [Current+SSH] & [Current] \\
            $4^{th}$ Experiment & [Current; SSH; Current+SSH] & [Current; SSH] \\
            \bottomrule
        \end{tabular}
    }
    \caption{Inputs and target variables for each experimental scenario we designed.}
    \label{tab:exp_design}
\end{table}

For all experiments we optimized our models with respect to the IoA \cite{Willmott1981}.
Hence, our loss function is defined as:
\begin{align}
    \resizebox{0.9\hsize}{!}{
        $\mathcal{L}(\hat{y_i},y_i) = 1 - {IoA} = 1 - \frac{\sum_{n=1}^N(y_i-\hat{y_i})}{\sum_{n=1}^N(\lvert \hat{y_i}-\bar{y_i} \rvert +\lvert y_i-\bar{y_i} \rvert)^2}$
    }
    \label{ioa}
\end{align}
\\
We also present results for Root-Mean-Squared Error (RMSE) in meters per second, and degrees.

\subsection{Model configurations}\label{subsec:model_configs}

For experiments, we used well-known frameworks like \textit{Pytorch} and \textit{Pytorch Geometric} \cite{Pytorch2019,PyGeometric2019} to implement the model, and \textit{Weights\&Biases} \cite{wandb} to track all experiments.

We used an LSTM as temporal encoder with 1 layer and a hidden dimension ($embed\_size$) of size 20, and 2 GNN blocks with GATv2 \cite{brody2021attentive} as GNN convolution. The only exception is the Transformers experiments, where we used 3 layers of Transformer Encoder with 5 attention heads, the same hidden dimension size and numbers of GNN blocks, but with a less complex message passage GNN convolution in contrast to GATv2, due to overfitting issues during training.

We ran each experiment with 10 different seeds. We here report the average performance of these 10 runs. Again, the exception is the Transformers experiments, which we ran for the best 5 seeds used with the LSTM as temporal encoder and report the average performance for these 5 runs.

The results for the first and the second experiments are reported in Table \ref{tab:current} and Table \ref{tab:elevation}. The third experiment is reported in Table \ref{tab:connectedvsdisconnected}, while the last experiment is reported in Table \ref{tab:curr_transformervslstm} and Table \ref{tab:elev_transformervslstm}.

\subsection{Discussion}\label{subsec:discussion}

\begin{figure}[htb]
    \centering
    \includegraphics[width=0.48\textwidth]{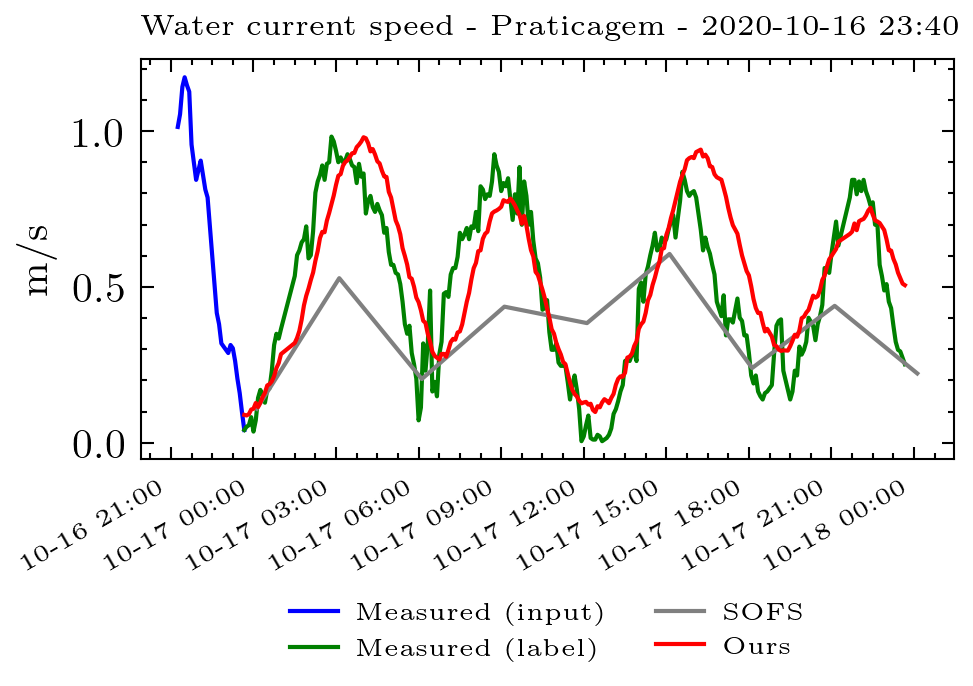}
    \caption{Comparison between our model and SOFS for current speed in meters per second for a sample $t_0$ in the test dataset. The full input window is 7 days long and is partially omitted for better visualization.}
    \label{fig:quali_example_1}
\end{figure}

Table \ref{tab:current} shows that our model surpasses SOFS in all scenarios by more than 17\% and 8\% for water current speed and direction respectively. This represents a considerable leap in quality of prediction that can be better visualized in Figure \ref{fig:quali_example_1}.
While results for all scenarios consistently showed that adding graph connectivity can benefit the model's performance, most improvement is still observed with a single variable as input.
This result indicate that the \emph{temporal encoder} is the major component in this gain.
Further indication of this can be found in Table \ref{tab:curr_transformervslstm} in which utilizing a Transformer architecture instead of an LSTM performed even better, surpassing SOFS by more than 27\% and 14\% for water current speed and direction, respectively.

Nevertheless, both Tables \ref{tab:current} and \ref{tab:curr_transformervslstm} show relevant gains to water velocity predictions when SSH nodes are added to $\mathcal{G}$. This shows that our model is able to aggregate information from SSH nodes into water current nodes' representation. SSH results in Table \ref{tab:elevation} and \ref{tab:elev_transformervslstm} did not benefit from the addition of current velocity data, but it is important to consider that SSH modeling in SOFS is already highly accurate and may present fewer opportunities of improvement. 

Another important effect to note is that adding nodes can decrease the model's performance. A small indication of this effect can be seen by a slight decrease in IoA when wind data is added. More directly, by analysing Table \ref{tab:connectedvsdisconnected} it is possible to note that a fully disconnected architecture performs slightly better than a graph with only nodes of the same \emph{type} connected in the case of water current velocity modeling. This indicates that aggregating current velocity data from other observation sites did not benefit our model's accuracy.

This effect may be related to both \emph{over-squashing} \cite{alon2021on,toppingUnderstandingOversquashingBottlenecks2021} of node embeddings and to the fact that our model uses a homogeneous GNN model in which functions $\phi$ and $\gamma$ from expression (\ref{mpnn}) are the same for all \emph{types}. Heterogeneous GNN models \cite{zhangHeterogeneousGraphNeural2019} have been demonstrated to perform well in scenarios with multiple node and edge types and constitute a future research direction for our work.

\begin{table}[htb]
    \centering
    \resizebox{\columnwidth}{!}{%
        \begin{tabular}{*{11}{l}}
            \toprule
                               & \multicolumn{2}{c}{\textbf{Speed (m/s)}} & \multicolumn{2}{c}{\textbf{Direction (degrees)}}                                      \\
            \cmidrule(lr){2-3}
            \cmidrule(lr){4-5}
            Scenarios             & IoA $\uparrow$                           & RMSE $\downarrow$                                & IoA $\uparrow$ & RMSE $\downarrow$ \\
    
            \midrule
            SOFS               & $0.599$                                  & $0.178$                                          & $0.755$        & $85.18$           \\
            Current            & $0.706$                                  & $0.165$                                          & $0.818$        & $70.25$           \\
            Current+SSH      & $\mathbf{0.726}$                                  & $0.158$                                          & $0.842$        & $65.68$           \\
            Current+SSH+Wind & $0.718$                                  & $0.160$                                          & $\mathbf{0.843}$        & $65.29$           \\
            \bottomrule
        \end{tabular}
    }
    \caption{Average results for 10 random seeds using data from all measuring stations to forecast Praticagem station water current speed and direction. In bold the best ones.}
    \label{tab:current}
\end{table}

\begin{table}[htb]
    \centering
    \begin{tabular}{*{11}{l}}
        \toprule
                      & \multicolumn{2}{c}{\textbf{SSH (m)}}                     \\
        \cmidrule(lr){2-3}
        \cmidrule(lr){4-5}
        Scenarios        & IoA $\uparrow$                                  & RMSE $\downarrow$ \\

        \midrule
        SOFS          & $0.935$                                         & $0.133$           \\
        SSH     & $\mathbf{0.940}$                                         & $0.124$           \\
        Current+SSH & $0.939$                                         & $\mathbf{0.123}$           \\
        \bottomrule
    \end{tabular}
    \caption{Average results for 10 random seeds using data from all measuring stations to forecast Praticagem SSH. In bold the best ones.}
    \label{tab:elevation}
\end{table}

\begin{table}[htb]
    \centering
    \resizebox{\columnwidth}{!}{%
        \begin{tabular}{*{11}{l}}
            \toprule
                               & \multicolumn{2}{c}{\textbf{Speed (m/s)}} & \multicolumn{2}{c}{\textbf{Direction (degrees)}}                                       \\
            \cmidrule(lr){2-3}
            \cmidrule(lr){4-5}
            Scenarios           & IoA $\uparrow$                           & RMSE  $\downarrow$                               & IoA $\uparrow$ & RMSE  $\downarrow$ \\
    
            \midrule
            Fully disconnected & $0.719$                                    & $0.161$                                            & $0.830$          & $68.89$              \\
            Same \emph{type} connections & $0.706$                                    & $0.165$                                            & $0.818$          & $70.25$              \\
            Fully connected    & $\mathbf{0.726}$                                    & $0.158$                                            & $\mathbf{0.842}$        & $65.68$               \\
            \bottomrule
        \end{tabular}
    }
    \caption{Average results for 10 random seeds comparing fully disconnected, same-variable connected and fully connected graphs in the [Current;SSH] predicting [Current] scenario. In bold the best ones.}
    \label{tab:connectedvsdisconnected}
\end{table}

\begin{table}[htb]
    \centering
    \resizebox{\columnwidth}{!}{%
        \begin{tabular}{*{11}{l}}
            \toprule
                          & \multicolumn{2}{c}{\textbf{Speed (m/s)}} & \multicolumn{2}{c}{\textbf{Direction (degrees)}}                                            \\
            \cmidrule(lr){2-3}
            \cmidrule(lr){4-5}
                          & \multicolumn{2}{c}{IoA}                  & \multicolumn{2}{c}{IoA}                                                                     \\
            \cmidrule(lr){2-3}
            \cmidrule(lr){4-5}
            Scenarios        & Transformer                              & LSTM                                             & Transformer             & LSTM           \\

            \midrule
            Current       & $\mathbf{0.734}$                  & $0.706$                                   & $\mathbf{0.841}$ & $0.818$ \\
            Current+SSH & $\mathbf{0.762}$                  & $0.726$                                   & $\mathbf{0.860}$ & $0.842$ \\
            \bottomrule
        \end{tabular}
    }
    \caption{Comparison of results for different time series encoders. In these experiments we used a Transformer, a state-of-the-art sequence model, as our first encoders, and compare its results with the previous ones, where we used an LSTM instead. Given the computational cost of Transformers, we used, for each input experiment, the 5 random seeds that gave the best results using LSTM as encoder, and averaged them. We used the same data as the previous experiments to forecast the speed and direction of the water current at Praticagem. In bold the best ones.}
    \label{tab:curr_transformervslstm}
\end{table}

\begin{table}[htb]
    \centering
    \begin{tabular}{*{11}{l}}
        \toprule
                      & \multicolumn{2}{c}{\textbf{SSH (m)}}\\
        \cmidrule(lr){2-3}
                      & \multicolumn{2}{c}{IoA}\\
        \cmidrule(lr){2-3}
        Scenarios & Transformer & LSTM\\

        \midrule
        SSH & $0.932$ & $0.940$ \\
        Current+SSH & $0.939$ & $0.939$ \\
        \bottomrule
    \end{tabular}
    \caption{Comparison of results for different time series encoders. Same experimental design as showed in Table \ref{tab:curr_transformervslstm}, except that we used SSH as solo input in one experiment, and we forecast the SSH at Praticagem.}
    \label{tab:elev_transformervslstm}
\end{table}

\begin{figure}[htb]
    \centering
    \includegraphics[width=0.48\textwidth]{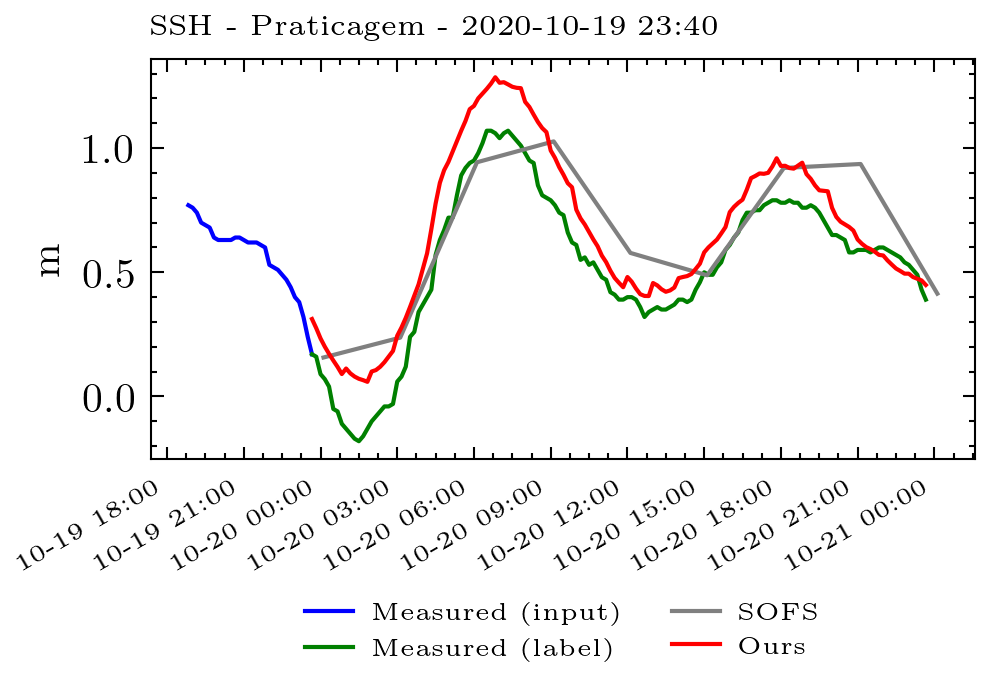}
    \caption{Comparison between our model and SOFS for SSH in meters for a sample $t_0$ in the test dataset. Values can be negative due to local SSH measurement offset.}
    \label{fig:quali_example_2}
\end{figure}

\section{Conclusion} \label{sec:conclusions}

We presented a real problem whose structure benefits from  a Graph Neural Network modeling strategy that encodes spatio-temporal features and their relations. Our experiments show the feasibility of a data-driven model capable of handling real datasets with missing data  and able to incorporate and share multivariate time series information between problem entities that are spatially separated. Furthermore, our model has better performance when compared with physics-based models for the task of forecasting oceanic variables.

In future work, we intend to look into the following directions:

\paragraph*{Latent graph inference:} When learning on graphs, one must assume the structure (topology) of the graph \textit{a priori}. However, any such hypothesis may be wrong; in large graph scenarios we end up with an enormous combinatorial problem. So, schemes such as advanced by \cite{kaziDifferentiableGraphModule2022} offer a promising direction to address larger graphs that may include data from other meteorological stations near SSVBES, increasing the number of nodes to hundreds or even thousands, while we learn the nodes connectivity on training.
\paragraph*{Heterogeneous GNNs:} Our experiments suggest that distinct oceanic variables have different levels of influence in forecasting the underlying target variables. Thus, we must search for interesting ways to handle heterogeneous graphs, where the nodes may be of different types, such as in \cite{schlichtkrullModelingRelationalData2018}. The heterogeneous setup may allow for different \emph{message functions} to be learned for different types of relations.
\paragraph*{Using SOFS as input:} SOFS considers other types of information such as local topology;  using that as a targetless node in the DTDG model may improve the performance even further. Recent proposals point in that direction within \textit{Physics-Informed Machine Learning} (PIML) \cite{Willard2020}.
\paragraph*{Temporal Graph Networks:} The proposals by \cite{rossiTemporalGraphNetworks2020} may also be
investigated   to encode temporal relationships between nodes, thus eliminating the need for the sampling process to form a DTDG.

\section*{Acknowledgements}

This work was carried out at the Center for Artificial Intelligence (C4AI-USP), with support by the Sao Paulo Research Foundation (FAPESP) under grant number 2019/07665-4 and by the IBM Corporation. This work is also supported in part by FAPESP under grant number 2020/16746-5, the Brazilian National Council for Scientific and Technological Development (CNPq) under
grant numbers 312180/2018-7, 310127/2020-3, 310085/2020-9, the Coordination for the Improvement of Higher Education Personnel (CAPES Finance Code 001), Brazil, and \textit{Ita\'{u} Unibanco S.A.} through the \textit{Programa de Bolsas Ita\'{u}} (PBI) program of the \textit{Centro de Ci\^{e}ncia de Dados} (C$^2$D) of \textit{Escola Polit\'{e}cnica} of USP. The authors also thank the Santos Marine Pilots for providing crucial data for this research.  

\bibliographystyle{named}
\bibliography{ijcai22}

\end{document}